\documentclass[a4paper,conference]{IEEEtran}
\ifCLASSINFOpdf
\else
\fi
\usepackage{times}
\usepackage{epsfig}
\usepackage{graphicx}
\usepackage{amsmath}
\usepackage{amssymb}
\usepackage{multirow}
\usepackage{subcaption}
\usepackage{xcolor}

\begin{document}
%
\title{ActionSpotter: Deep Reinforcement Learning Framework for Temporal Action Spotting in Videos}

\author{\IEEEauthorblockN{Guillaume Vaudaux-Ruth}
\IEEEauthorblockA{ONERA, Sorbonne Universit\'e\\
Palaiseau, France\\
guillaume.vaudaux-ruth@onera.fr}
\and
\IEEEauthorblockN{Adrien Chan-Hon-Tong}
\IEEEauthorblockA{ONERA, Universit\'e Paris Saclay \\
Palaiseau, France\\
adrien.chan\_hon\_tong@onera.fr}
\and
\IEEEauthorblockN{Catherine Achard}
\IEEEauthorblockA{Sorbonne Universit\'e, CNRS UMR 7222, ISIR\\ F-75005, Paris, France\\
catherine.achard@sorbonne-universite.fr}}


%


\maketitle



%
\IEEEpeerreviewmaketitle

\begin{abstract}
Action spotting has recently been proposed as an alternative to action detection and key frame extraction.
However, the current state-of-the-art method of action spotting requires an expensive ground truth composed of the search sequences employed by human annotators spotting actions - a critical limitation.


In this article, we propose to use a reinforcement learning algorithm to perform efficient action spotting using only the temporal segments from the action detection annotations, thus opening an interesting solution for video understanding.

Experiments performed on THUMOS14 and ActivityNet datasets show that the proposed method, named ActionSpotter, leads to good results and outperforms state-of-the-art detection outputs redrawn for this application.
In particular, the spotting mean Average Precision on THUMOS14 is significantly improved from 59.7\% to 65.6\% while skipping 23\% of video.
\end{abstract}

\section{Introduction}
Many works are interested in the analysis of actions in videos as it leads to a lot of applications. 
For example, it can be used to index Youtube videos \cite{gu2018ava} or to control robots by gestures \cite{rautaray2015vision}. 
The evaluation of the quality of actions \cite{morel2017automatic} can be used to improve sports performances and the segmentation of video streams is well suited to monitor elderly people in their homes \cite{chan2014simultaneous}.
Action detection \cite{yeung2016end} is also an interesting subject since it aims at detecting action realizations in untrimmed videos. Thus, in addition to the action class, action detection returns the precise start and end times of each action instance. It can be used for video surveillance by detecting accidents, thefts or fights, but also for people protection (e.g. detection of falls).

The visual summary of videos has also been widely studied in particular to provide a smart scroll bar when streaming videos. The two most common frameworks are key frame selection \cite{liu2003novel,kuanar2013video,ejaz2013efficient,zhang2016video} and key sub-shot selection \cite{mademlis2016movie,mademlis2018salient}. Both frameworks are classically unsupervised and aim at finding clusters that best describe all frames. 

\begin{figure}[t]
\includegraphics[width=0.45\textwidth]{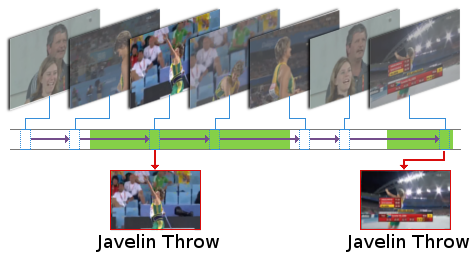}
\caption{\textbf{Overview of the proposed spot frame extraction framework.} It extracts one spot frame per action occurrence (in green in the figure) by sparsely browsing the video. 
\label{fig:overview1}}
\end{figure}


However, to summarize the actions in videos, these techniques are not optimal. Indeed, it can be very difficult to find precise action boundaries, as it is done in action detection. Furthermore these boundaries are not necessary to know the existence of an action. Moreover, finding a visual summarizing does not necessarily reflect the number of actions. In fact, the visual summarizing is intended to extract salient or key images from a video stream. Thus, an action composed of several visual parts will be described using several key frames, which does not allow to count the instances of actions. This issue also prohibits reliance on single-image action classification \cite{zhao2017single,guo2014survey} for such task, as this could lead to discontinuous segments and subsequent overestimation of action occurrences.

To overcome these limitations, action spotting has recently been proposed in \cite{alwassel2018action}: it consists in producing an ordered list of the action instances contained in the video, or in other words \textit{"the process of finding any temporal occurrence of an action"} as defined in \cite{alwassel2018action}. 
Typically, it can be done by selecting one frame per action instance, as shown in figure~\ref{fig:overview1}.

To the best of our knowledge, the spotting task has only recently appeared in the literature: Alwassel \textit{et al.} \cite{alwassel2018action} use it as pre-processing for detection, Bhardwaj \textit{et al.} \cite{bhardwaj2019efficient} or Wu \textit{et al.} \cite{wu2019adaframe} use it to produce accurate classification. 

In this work, we consider action spotting as an individual task (and not to speed up classification or improve detection). The resulting spot frames are not expected to be key frames, and are not annotated in the ground truth (only composed of action segments): they are automatically selected by the method to optimize the spotting results. 
Thus, for a given action segment, the selection of any frame belonging to this time interval as a spot frame is correct. Consequently, two spotting outputs can be different but equally accurate. 
The prediction is considered perfect if and only if there is exactly one spot frame per ground truth segment, with a correct label (allowing for example a perfect count).
Thus, the problem is asymmetric: while the ground truth is composed of a set of segments, the prediction is a set of spot frames.
We therefore introduce a metric to directly measure the quality of a set of spot frames.
It is based on the mean Average Precision (mAP) value used in the AVA dataset toolbox \cite{gu2018ava} adapted to action spotting.
The above-mentioned measure, which estimates spotting performance, is available in supplementary material and is described in more detail in section \ref{sec:metric}. 
In addition to defining this new metric, we have also redrawn the predictions of state-of-the-art action detectors to fit the proposed metric, as will be detailed in section \ref{sec:Comparison}

Independently, in order to carry out fully supervised action spotting, \cite{alwassel2018action} introduced expensive annotations on the behavior of human annotators when performing action spotting. We believe that supervised learning is not straightforward as the problem is asymmetric, and we propose the use of reinforcement learning to produce spot frames from action segments only. This algorithm, called ActionSpotter, can handle videos containing multiple classes of action with multiple occurrences and is described in section \ref{sec:method}.

To summarize, our contributions are as follows:
\begin{itemize}
    \item We release a metric for the spotting task (see attachment).
    \item We propose to use deep reinforcement learning algorithm to avoid the time-consuming annotations of human behavior and we propose to spot frames in a semi-supervised way (only action segments are available during training).
    \item We propose ActionSpotter, a reinforcement learning architecture extracting spot frames while observing as few frames as possible. It is based on
    the state-of-the-art actor-critic architecture. 
    \item We show that this architecture is more relevant for action spotting than a post-processing of state-of-the-art detectors and/or spotting baselines.
\end{itemize}
Our framework is presented in Section \ref{sec:method_section} after related works in Section \ref{sec:related_works}. Then, experiments on THUMOS14 and ActivityNet are presented in Section \ref{sec:experiments} before conclusion.

\begin{figure*}[t]
\centering
\includegraphics[width=0.9\textwidth]{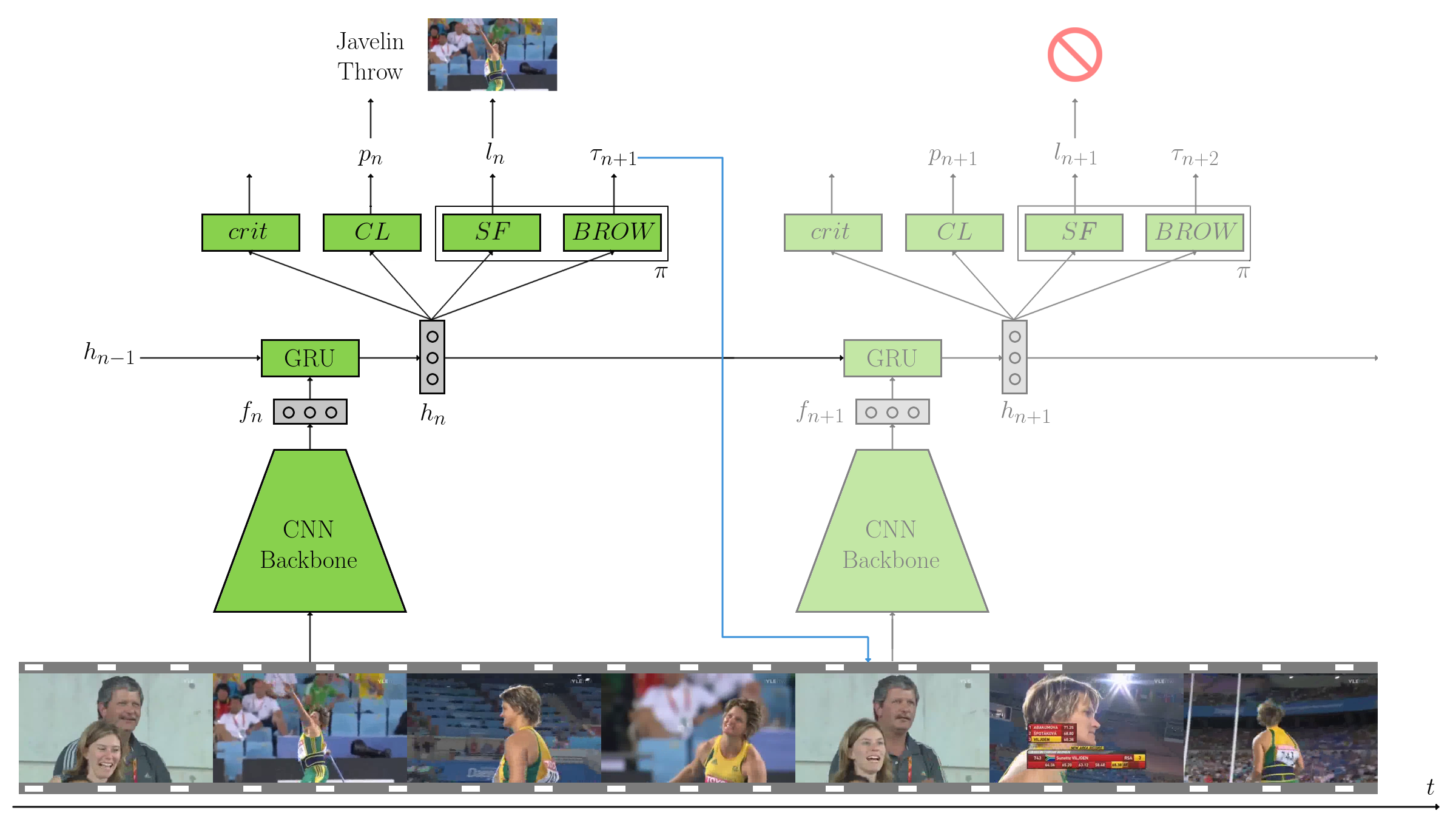}
\caption{\textbf{Overall description of our pipeline.} \label{fig:image3} In a first stage, the CNN backbone encodes the frame (or non-overlapping chunk of frames) into a feature vector which is then forwarded to a GRU layer. The resulting hidden state vector is then individually processed by (SF), (CL), (BROW) and ($crit$). The (SF) stage deals with the decision of turning the current frame into a spot frame or to skip it. The (CL) stage predicts the action class related to the spot frame and the (BROW) stage outputs the next video frame to look at. The (crit) stage is only used to ensure better convergence in the reinforcement learning framework.
}
\end{figure*}

\section{Related works}
\label{sec:related_works}
Algorithms based on deep reinforcement learning have been proposed in literature to quickly browse videos, especially for fast video classification \cite{wu2019adaframe,wu2019multi}, early video detection \cite{ma2016learning,gao2019startnet} or action detection \cite{yeung2016end,alwassel2018action}.

For example, \cite{wu2019adaframe} offers an impressive result producing a state-of-the-art classifier by exploring only (on average) 8 frames per video.
However, this is only possible thanks to the prior that videos contain a single action (or even object): in this context, a single good image could be enough to decide the class of the whole video.

In detection or spotting, as this assumption does not hold, performance decreases quickly when frames are skipped. This is particularly evident for short actions in long videos that can be completely missed by a greedy browsing.
Typically, Giancola \textit{et al.} \cite{Giancola2018SoccerNetAS} introduced a benchmark for locating very sparse events in soccer videos and propose a method based on a classification of video slices. However, as actions are very short, they process all the frames.
Only \cite{yeung2016end} tackles detection using few images. 

In this paper, we rely on reinforcement learning, but for another reason: spot images are not provided by the ground truth, which is only composed of temporal segments.
Thus, even though \cite{wu2019adaframe} also uses an architecture based on reinforcement learning, the goal of our work is different, our objective being to produce a good spotting. Such an idea can be found  for object spotting \cite{perreault2020spotnet} in 2D images, but we believe we are the first to apply it to temporal action spotting. Our framework also does not require video-retargeting \cite{bansal2018recycle,wolf2007non}.

Moreover, our framework balances the trade-off between accuracy and frames skipping and, in opposition to  \cite{yeung2016end,wu2019adaframe,wu2019multi,alwassel2018action}, ActionSpotter processes videos online. Indeed, we show that using the current frame and a memory of previous ones is sufficient to take correct decisions.


These considerations lead to a practical spotting algorithm that does not rely on detection or segmentation. This property is interesting as detectors have to tackle a much more difficult task as they have to find the starting and ending points of actions that are often ambiguously defined.

The article closest to our work is \cite{alwassel2018action} where Alwassel \textit{et al.} propose an algorithm that produces one spot per video. Their process is fully supervised and is done by learning the trajectories made by a human during the exploration of the videos.
The authors argue that the use of supervised trajectories during training is much more direct and simple than reinforcement learning. 
However, this is an important limitation of this method as it requires a lot of human acquisition to obtain the browsing strategies of the videos. 
Moreover, \cite{alwassel2018action} does not consider the spot frame extraction for itself but only as a first step for action detection and thus does not present spotting results. 




\section{Method}
\label{sec:method_section}
\subsection{Action spotting and proposed evaluation metric}
\label{sec:metric}
In this section, we present the proposed methodology for action spotting. As presented in introduction, the goal is to browse a video in order to select \textit{spot frames} summarizing human activity in videos (see Figure \ref{fig:overview1}).
We are thus interested in optimizing the quality of spot frames (described below), but also the proportion of skipped frames, called the \textit{skip ratio}.

\textbf{Spot Frames:} If $C$ is a predefined set of action classes, and $V = \{ v_t\}_{t=1}^{T}$ a video sequence of $T$ frames (or frame chunks), containing a set of $K$ temporaly ordered actions proposals of interest $Q = \{q_k \mid q_k =(a_k, [I_k,J_k])\}_{k=1}^{K}$, with $a_k \in C$ the action class, and with $I_k$ and $J_k$ the start and end time of a proposal,
then our goal is to produce a set of spot frames with likelihoods and labels $\mathcal{V} = \{(\tau_k,l_k, \alpha_k)\}_{k=1}^K$ such that $\forall k\in \{1,...,K\}, \ \tau_k\in [I_k,J_k]$ and $\alpha_k = a_k$. 

\textbf{Evaluation Metric:} In order to propose an unbiased metric reflecting the quality of the extracted spot frames, we build it on the basis of the state-of-the-art metric used in object detection \cite{Everingham10} or temporal action localization \cite{caba2015activitynet}: the mean Average Precision (mAP). This metric is for example used on Pascal, THUMOS or AVA challenge.
Thus, we adapt this metric to action spotting and propose a new evaluation script publicly available to ensure a fair evaluation.


To compute this metric, for each predicted class, spot frames are sorted in decreasing order according  to  their likelihood. Then,  the  intersection between  the  timestamps  of  the  spot  frames  and  the  ground truth segments are computed iteratively. A spot frame is then flagged  as  a  correct  detection  if  and  only  if  its  time stamp intersects a ground truth segment, is classified with the correct label and is the first to match with the ground truth segment. A  spot  frame  that  does  not  match  with  any  ground  truth segment  or  is  not  the  first  to  match  with,  is  a  false  alarm. Finally,  a  ground  truth  segment  that  does  not  matched  with any  spot  frame  corresponds  to  a  missed  detection. Then, for each action class, we compute precision and recall values at every position in the ranked sequence of spot frames, in order to plot the precision-recall curve.
Finally, the areas under these curves are averaged over all activity classes and this mean value is returned as the quality of the spotting (mAP).

To the best of our knowledge, we are the first to focus on the quality of a set of spot frames for itself.

\subsection{ActionSpotter: Actor Critic based semantic spot frame extractor}
\label{sec:method}
We designed a pipeline called ActionSpotter containing three networks that work together to both browse video frames in an online way and extract spot frames reflecting human activity. 
The overall state of our pipeline at timestamp $n$ is given by three elements: a current frame $\tau_n$, a memory $h_n$ and a set of spot frames/likelihoods/labels $\mathcal{V}_{n}$.
Importantly, frames are used following the video stream and can only be used once.

\textbf{Memory:} At time step $n$, the frame (or frame chunk) $v_{\tau_n}$ is forwarded to a backbone network $BB$, based on CNN. $BB$ extracts a feature vector $f_n$ that contains spatial information $f_n = BB(v_{\tau_n})$. Any state-of-the-art backbone network can be employed to implement $BB$.
Then, a Gated Recurrent Layer (GRU) \cite{chung2014empirical} is used to encode temporal information. It takes as inputs the feature vector $f_n$ and the previous hidden state $h_{n-1}$ and produces the current hidden state $h_n$, seen as a latent vector that contains the memory of the past viewed frames: $h_n = \text{GRU}(f_n, h_{n-1})$.

\textbf{Classification Network:} Then, the classification network $CL$ reads the current hidden state $h_n$ and produces a probability distribution over action classes $p_n= CL(h_n) \in \mathbb{R}^C$. The predicted action label is then such as $\alpha_n =  \underset{c}{\arg\max}(p_{n,c})$.

\textbf{Spot Frame Selector Agent:} The memory $h_n$ is also forwarded to the Spot Frame Selector Agent $SF$ which produces a likelihood for the current frame to be a good spot frame: $l_n = SF(h_n)$.
Formally, the output of our algorithm is updated as $\mathcal{V}_{n+1} = \mathcal{V}_{n} \cup \{\tau_n,l_n, \alpha_{n}\}$.
In production, one may select a detection threshold $\sigma$ keeping only spot frames with a likelihood greater than $\sigma$. This leads to a functioning point described by its precision and recall. 

\textbf{Browser Agent:} In parallel, this memory $h_n$ is also forwarded to the Browser Agent $BROW$ which decides the next frame to visit, i.e. $\tau_{n+1} = \tau_n + BROW(h_n)$.

\textbf{Skip ratio:} Importantly, $\tau_n\neq t$ as our pipeline does not go through all the images: let $N$ such that $\tau_N=T$ (i.e. the number of steps to process the video), then, the ratio of skipped frames is defined as 1-$\frac{N}{T}$.

\underline{\textbf{Global dynamic:}} Then, at step $n$, ActionSpotter ($AS$), has the following dynamic:
\begin{equation}\label{eq:1}
    AS(v_{\tau(n)}, h_{n-1}): \left\{ \begin{array}{c}
    f_n= BB(v_{\tau(n)}) \\
         h_n = \text{GRU}(f_n, h_{n-1}) \\
         l_n = SF(h_n) \\
         p_{n}= CL(h_n)\\
         \alpha_{n} = \underset{c}{\arg\max} \ (p_{n,c}) \\
         \mathcal{V}_{n+1} =
            \mathcal{V}_{n} \cup \{(\tau_n,l_n, \alpha_{n})\} \\
         \tau_{n+1} = \tau_n + BROW(h_n)  
    \end{array}  \right.
\end{equation}
with $h_{n-1}=\text{GRU}(BB(\{v_{\tau(i)}\}_{i=1}^{n-1}))$, the memory of the past viewed frames (or frame chunks).

Figure \ref{fig:image3} gives an illustration of the overall framework.
 
\subsection{Training and objectives}
Knowing that we use a pretrained model for $BB$ (see section \ref{sec:experiments}), the training goal is to optimize the parameters of GRU, $BROW$, $SF$ and $CL$ so that by processing the video $V$, the policy $\pi = BROW + SF$ provides an accurate set of spot frames $\mathcal{V}$, while skipping a large number of frames (i.e. with $BROW(h_n)$ as high as possible).
It is important to note that the browser $BROW$ and spot frame selection network $SF$ are not learned in a supervised way, removing the need for specific annotations.
Instead, a state-of-the-art reinforcement learning algorithm is used with a specific reward function quantifying the relevance of each step under policy $\pi$ (both for accuracy and browsing ratio).

\textbf{Reward.} As we use mAP to evaluate model performance, we choose to introduce a reward directly linked with this metric: the global policy has to maximize the final mAP of the video being processed (plus an entropy term  $\rho \mathcal{H}(\pi(n))$ which will be explained later).
Moreover, we can easily find a trade-off between the efficiency of video browsing and the accuracy of spot frame selection by discounting the final mAP by $\gamma^{N}$ where $\gamma \in (0,1]$ is the discount factor and $N$ the explored frames.

According to reward shaping theory \cite{laud2004theory}, this final reward can be hybridized with a potential-based shaping to help reinforcement convergence without changing optimal policy. Straightforwardly, using the mAP after each step is an interesting shaping potential.
Thus, our local reward $r_{\pi,n}$ at step $n$ is just the difference of mAP between step $n$ and $n-1$ under policy $\pi$ (plus entropy term):
\begin{equation}
   r_{\pi,n} = \gamma\text{mAP}(\mathcal{V}_{n}) - \text{mAP}(\mathcal{V}_{n-1}) + \rho \mathcal{H}(\pi(n))\label{eq:2}
\end{equation}
Then, $R_{\pi,n}$ the cumulative discounted rewards is:
\begin{equation}
R_{\pi,n} = \sum_{k=1}^{n}\gamma^k r_{\pi, k}
\end{equation}
By omitting entropy, we can directly verify that: 
\begin{align*}
R_{\pi,N} &= \sum_{k=1}^{N}\gamma^k r_{\pi, k} = \sum_{k=1}^{N}\gamma^k \gamma\text{mAP}(\mathcal{V}_{k}) - \gamma^k\text{mAP}(\mathcal{V}_{k-1}) \\
&= \gamma^{N+1}\text{mAP}(\mathcal{V}_{N}) - \text{mAP}(\mathcal{V}_{0})
\end{align*}
exactly as desired as $\text{mAP}(\mathcal{V}_{0})=0$. This invariance in shaping is true even with entropy as shown by \cite{laud2004theory}. 

\textbf{Actor-Critic optimization.} The Policy Gradient method is based on the total expected reward, and therefore requires a long sequence of actions to update the policy. Good and bad actions are then averaged, which can introduce convergence issues. Actor-Critic approach \cite{haarnoja2018soft} is known to be a way to avoid this issue by evaluating each action independently. It uses two models named actor and critic. 

The actor is straightforwardly trained to find the policy that maximizes the expected return:
\begin{equation}
J_{actor} = \mathbb{E}[R_{\pi, N}]
\end{equation}

The critic measures how good the policy is (value-based) and  produces an estimation of the value function which is the expected discounted reward $crit(h_n) \approx \mathbb{E}[R_{\pi,N}- R_{\pi,n} | h_n]$. Thus,
the loss function linked to the critic is defined as: 
\begin{equation}
\mathcal{L}_{critic} = \frac{1}{2} \|crit(h_n) - \mathbb{E}[R_{\pi,N}- R_{\pi,n} | h_n]\|_2
\end{equation}

In reinforcement learning, it is crucial to balance exploration and exploitation. Following the work of Haarnoja \textit{et al.} \cite{haarnoja2018soft}, it is possible to integrate this balance directly into the reward, by adding an entropy penalty which forces the actor to uniformly explore states with equal rewards.
Thus, a penalty $\rho \mathcal{H}(\pi(n))$ is added in equation \ref{eq:2}.
$\rho$ is the temperature parameter which balances exploration and exploitation and is automatically adjusted following \cite{haarnoja2018soft}. $\mathcal{H}()$ is the entropy function. 
Let us denote that without shaping or temperature, reinforcement has difficulty to converge.

In our algorithm, the actor is the combination of the Browser Agent $BROW$ and the Spot Frame Selector Agent $SF$. Thus, $\mathcal{H}(\pi(n))$ is the entropy applied  to the distribution $\pi(n)$ over choices to update the current state \textit{i.e.} $BROW(h_n)$ and $SF(h_n)$.

On the other side, the classification network CL is trained, in a supervised way, using Cross-Entropy (CE). Thus,
\begin{equation}
    L_{cls} = CE(p_n,a_{\tau_n})
\end{equation} 

\underline{\textbf{Final loss:}}
Combining previous losses, our final objective is to minimize:
\begin{equation}
\mathcal{L}_{global} = \mathcal{L}_{cls} + \lambda_1 \mathcal{L}_{critic} - \lambda_2 J_{actor}
\end{equation}

As $J_{actor}$ objective is non differentiable we use REINFORCE \cite{Sutton:1998:IRL:551283} to derive the expected gradient: 
\begin{equation}\label{eq:8}
     \nabla J_{actor} =\nabla \mathbb{E} \left[\sum_{n=1}^{N} \log (\pi(n)) (R_{\pi,n} - \mathbb{E}[R_{\pi,n} | h_n]) \right]
\end{equation}
We can then approximate this equation by using Monte Carlo sampling and finally use stochastic gradient descent to minimize our final objective.

\begin{table*}[t]
\begin{center}
\begin{tabular}{|l|c|c|c|c|c|c|}
 \multicolumn{7}{c}{}\\
\hline
 \multicolumn{7}{|c|}{\textbf{THUMOS'14}}\\\cline{1-7}
 \multirow{2}{*}{Approach} & \multicolumn{5}{c|}{Detection mAP@} & \multirow{2}{*}{Spotting mAP}\\\cline{2-6}
 & 0.1 & 0.2 & 0.3 & 0.4 & 0.5 & \\\hline
\hline
Glimpses \cite{yeung2016end}& 48.9 & 44.0 & 36.0 & 26.4 & 17.1 & - \\
SMS \cite{yuan2016temporal}& 51.0 & 45.2 & 36.5 & 27.8 & 17.8 & - \\
M-CNN \cite{scnn_shou_wang_chang_cvpr16}& 47.7 & 43.5 & 36.3 & 28.7 & 19.0 & 41.2\\
CDC \cite{cdc_shou_cvpr17} & - & - & 41.3 & 30.7 & 24.7 & 31.5\\ 
TURN \cite{Gao_2017_ICCV} & 54.0 & 50.9 & 44.1 & 34.9 & 25.6 & 44.8\\
R-C3D \cite{DBLP:journals/corr/XuDS17}& 54.5 & 51.5 & 44.8 & 35.6 & 28.9 & 52.2\\
SSN \cite{zhao2017temporal}& \textit{66.0} & \textit{59.4} & 51.9 & 41.0 & 29.8 & - \\
A-Search \cite{alwassel2018action}& - & - & 51.8 & 42.4 & 30.8 & - \\
CBR \cite{gao2017cascaded}& 60.1 & 56.7 & 50.1 & 41.3 & 31.0 & 50.1\\
BSN + UNet \cite{DBLP:journals/corr/abs-1806-02964} & - & - & 53.5 & 45.0 & 36.9 & -\\
Re-thinking F-RCNN \cite{47058} & 59.8 & 57.1 & 53.2 & 48.5 & 42.5 & -\\
D-SSAD \cite{DBLP:journals/corr/abs-1904-07442} & - & - & \textit{60.2} & \textit{54.1} & \textit{44.2} & \textit{59.7}\\\hline
Ours (TSN backbone) & -  & -  & -  & - & - & 62.4\\
Ours (I3D backbone) & -  & -  & -  & - & - & \textbf{65.6}\\\hline

\end{tabular}
\end{center}
\caption{\label{tab:main} \textbf{Results on THUMOS14 validation set.} Second column: state-of-the-art detector results according to detection metric, computed with different IOU thresholds ranking from $0.1$ to $0.5$. Last column: mAP results for the spotting task. 
}
\end{table*}

\section{Experiments}
\label{sec:experiments}
Currently, we are the first to tackle action spotting as an individual task. It is therefore not easy to position our algorithm regarding the start-of-the-art. So, we offer two experiments to highlight the capabilities of this algorithm.
First, we show that ActionSpotter produces better sets of spot frames than post-processed state-of-the-art detectors. Then, we compare ActionSpotter to several baselines to highlight that action spotting is a challenging task and that reinforcement learning is the relevant way to deal with this asymmetrical problem where the ideal output is not defined in the ground truth.
Finally, some additional experiments reveal that ActionSpotter is able to balance accuracy and skip ratio, simply by modifying the discount factor associated with reinforcement learning.

\subsection{Datasets}
We evaluate our approach on the well-known THUMOS14 \cite{THUMOS14} and ActivityNet \cite{7298698} datasets.

\textbf{THUMOS14} dataset is composed of 101 activity classes for action recognition and a subset of 20 classes for action detection. 
Validation and testing sets contain respectively 200 and 212 untrimmed videos temporally annotated with action instances. 
We adopt the classical train/test setting of THUMOS14 protocol: training is done on 20 classes validation set and evaluation is done on testing set - original training set being not suited for detection.

\textbf{ActivityNetv1.2} dataset contains 9,682 videos in 100 classes collected from YouTube. The dataset is divided into three subsets: 4,819 videos for training, 2,383 for validation and 2,480 for testing. Action spotting results on ActivityNet dataset are reported on validation set as evaluation server does not compute our spotting metric.

\subsection{Implementation details} 
As previously mentioned, any type of backbone network can be used to encode local information from images. In order to have the same backbones as the state-of-the-art action detectors, we rely on classical TSN \cite{NIPS2014_5353} and I3D \cite{DBLP:journals/corr/CarreiraZ17} feature extractors. This setting allows good reproducibility as such features are provided by \cite{DBLP:journals/corr/abs-1807-10418} and \cite{DBLP:journals/corr/abs-1806-02964}. These techniques operates on both RGB frames and optical-flow field to capture appearance feature and motion information. TSN operates on individual video frames while I3D features are extracted from non-overlapping 16-frame video slices. 
For this second technique, as feature represents non-overlapping frame slices, the Browser Agent BROW does not process individual frames but frame slices. This does not change the skip ratio because it is equivalent to considering one slice in $K$ instead of one frame in $K$.

For the memory, we use a one-layer GRU with respectively 2,048 and 400 hidden units for THUMOS14 and ActivityNet (the size of the respected features). BROW, SF, CL and $crit$ have 3 linear layers and ReLu activation function. The first two layers have the same number of hidden units as the GRU layer and the last one has the size of the network output.

In our main setting, BROW can choose to move to the next frame, skip one frame or  skip three frames. At training time, in order to approximate Eq.\ref{eq:8}, the actions performed by BROW and SF are sampled from a categorical distribution parameterized by their respective logits. At testing time, the action performed is the one with the highest likelihood for BROW (and for CL). SF directly outputs a likelihood.

We use PyTorch for implementation and Adam for optimization with an initial learning rate of $10^{-4}$ and a batch size of 32.
Convergence is much faster when $CL$ and $SF$ are trained alone for few epochs before starting the whole reinforcement learning process with $\lambda_1 = \lambda_2 = 1$.

\subsection{Comparison between ActionSpotter and detectors}
\label{sec:Comparison}

\begin{table}[h]
\begin{center}
\begin{tabular}{|l|c|c|c|c|c|}
\multicolumn{6}{c}{}\\
\hline
 \multicolumn{6}{|c|}{\textbf{ActivityNet v1.2}}\\\cline{1-6}
\multirow{2}{*}{Approach}& \multicolumn{4}{c|}{Detection mAP@}&\multirow{2}{*}{Spotting mAP}\\\cline{2-5}
 & 0.5& 0.75 & 0.95 & Avg &\\
\hline\hline
W-TALC \cite{paul2018w} & 37.0 & 14.6 & - & 18.0& -\\
SSN-SW \cite{zhao2017temporal}  & - & - & - & 18.1& -\\
3C-Net \cite{narayan20193cnet}  & 37.2 & 23.7 & 9.2 & 21.7& -\\
FPTADC \cite{inbook} & 37.6 & 21.8 & 2.4 & 21.9& -\\
SSN-TAG \cite{zhao2017temporal} & 39.2 & 25.3 & 5.4 & 25.9 & 55.4\\
BSN \cite{lin2018bsn} & 46.5 & 30.0 & 8.0 & 30.0 & 49.6 \\
BMN \cite{lin2019bmn} & 50.1 & 34.8 & 8.3 & 33.85 & 55.3 \\\hline
Ours (TSN backbone) & -  & -  & -  & -  & 58.1 \\ 
Ours (I3D backbone) & -  & -  & -  & -  & \textbf{60.2} \\ 
\hline
\end{tabular}
\end{center}
\caption{\label{tab:activitynet} \textbf{Results on ActivityNet v1.2 validation set.} The column AVG indicates the average mAP for IoU thresholds: 0.5:0.05:0.95.}
\end{table}

We evaluate ActionSpotter performances on \textbf{THUMOS14} and \textbf{ActivityNet} datasets using our proposed spotting metric. Results are presented in Table~\ref{tab:main} and \ref{tab:activitynet}.
These tables also present results obtained by state-of-the-art detectors for the spotting task.
We use published results or available codes to obtain detection results, so there is no re-implementation issues. Detection results are redrawn into spotting results by extracting the centers of the predicted segments.

These experiments show that ActionSpotter significantly outperforms state-of-the-art detectors for the spotting task: the mAP of the latest action detector D-SSAD \cite{huang2019decoupling} is improved from 59\% to 65\% on THUMOS14 database.

It can be pointed out that FrameGlimpses \cite{yeung2016end} offers very low performance but has a skip ratio of 98\%. Currently, ActionSpotter performance also decreases from 62.4\% to 50.9\% when using TSN backbone and increasing the skip ratio to 98\%.
This shows that it is difficult to do accurate spotting (and even more detection) with only 2\% of the frames, which is not the case for action classification \cite{wu2019adaframe}.

One may wonder why spotting performance are not similar to detection performance with a low Intersection over Union (IoU) value. In fact, this is due to a difference in the matching mechanism. 
In the case of detection, predicted segments match with ground-truth segments according to the best IoU while, in the case of spotting, the predicted spots match the ground-truth according to their scores (which make sense since IoU no longer exists).
Thus, for detection task, it is better to predict segments with good localization and random confidence than to produce one segment per action with good confidence but coarse localization. Conversely, only the confidence score matters in spotting.
This difference between the matching processes induces some changes in the ranking of predictions between detection and spotting: while CDC \cite{cdc_shou_cvpr17} is more efficient in detection than M-CNN\cite{scnn_shou_wang_chang_cvpr16}, it is the opposite for spotting.

Thus, spotting results obtained with detectors cannot be easily compared with those of spotting algorithms as their primary goal is not the same. However, the large performance gap between our method and detection methods shows that it is not sufficient to post-process detector results to have an optimal spotting, and that specific algorithms such as ActionSpotter are required.
 
\subsection{Ablation study and comparison with baselines}
Previous experiments show that action spotting requires a specific approach. Moreover, action spotting is an asymmetric problem since any frame of a ground truth segment can be used as a spot frame and, therefore, optimal spot frames are not defined by these segments.
Based on this, Reinforcement Learning appears to be suited to tackle this problem and we show: (\textit{i}) comparative results between supervised and reinforced spotting algorithm, (\textit{ii}) an ablated version of ActionSpotter. Results are presented in table \ref{tab:ablation}. 

\begin{table}[h]
\begin{center}
\begin{tabular}{|l|c|}
\multicolumn{2}{c}{}\\
\hline
Method & mAP (\%)\\
\hline\hline
Naive Segmentation & 32\\
Multi-Task Segmentation & 43\\
Supervised ActionSpotter  & 52\\ 
No Memory ActionSpotter  & 45\\ 
\hline
\textbf{ActionSpotter} (memory + reinforcement) & \textbf{65}\\
\hline

\end{tabular}
\end{center}
\caption{\label{tab:ablation} \textbf{Comparison between ActionSpotter and other spotting algorithms on THUMOS14 validadation set.
}
}
\end{table}

Naive segmentation is a simple semantic CNN-based segmentation  (based on I3D + 2 layers)  where the ground truth is constructed as follows: the center of each action segment is labeled by its action class and all other frames are labeled as background.
This spotting-oriented baseline, only reaches 32\% of mAP.

A two-tasks learning process is then used as a multi-task segmentation: the first task learns the center of segments and the second predicts the action class. This second baseline deals with a balanced problem helping gradient stabilization and leads to 43\% of mAP.

\begin{figure*}[h]
 
\begin{subfigure}{0.33\textwidth}
\includegraphics[width=1\linewidth, height=3.2cm]{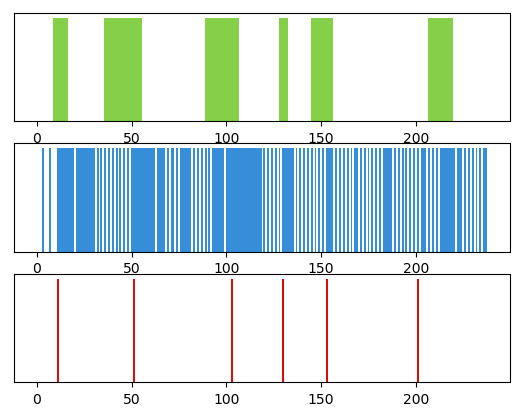} 
\label{fig:subim1}
\end{subfigure}
\begin{subfigure}{0.33\textwidth}
\includegraphics[width=1\linewidth, height=3.2cm]{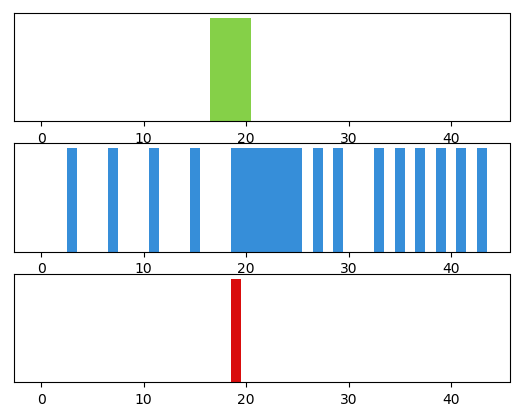}
\label{fig:subim2}
\end{subfigure}
\begin{subfigure}{0.32\textwidth}
\includegraphics[width=1\linewidth, height=3.2cm]{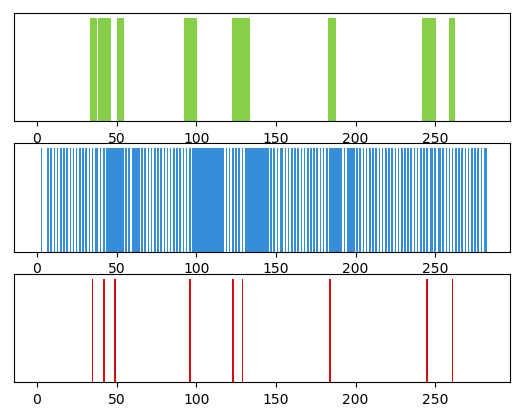}
\label{fig:subim3}
\end{subfigure}
 
\caption{\label{fig:image2} \textbf{Example of ActionSpotter outputs.} The figure displays outputs of ActionSpotter on THUMOS14 validation videos: columns represent the frames and, the 3 rows represent respectively ground truth segments, explored frames and selected spot frames. The same results are presented in video in supplementary material.}
\end{figure*}

Then, we train the ActionSpotter architecture to the supervised segmentation of action centers (instead of reinforcement learning).
In this setting, performance reaches 52.3\% mAP, which is higher than other baselines but well below ActionSpotter.

Finally, we consider ActionSpotter (with reinforcement learning) by removing its memory. Performance drops to 45\%: as expected, without memory, reinforcement and supervision are practically the same since the decision is frame based.

These results highlight the fact that it is not easy to supervise action spotting.  Actually, it is preferable to let the network select the easiest spot frame rather than imposing them. That's exactly what reinforcement does as the reward is only based on the final output. 

Figure~\ref{fig:image2} shows qualitative results on three videos: in most cases, the spot frames (third row in red) are not extracted in the middle of the actions (first row in green).
ActionSpotter adapts itself to the difficulty of the processed videos, as shown by the fact that the browser explores less frames (second row in blue) in areas without actions.
These results are presented in video in supplementary material.

In addition, ActionSpotter allows direct optimization of the mAP score using state-of-the-art shaping technique.
Indeed, this shaping technique is an important component as performance drops to 47\% without it.
ActionSpotter is therefore based on 3 key ideas: end-to-end learning, reinforcement learning to tackle asymmetry and shaping to allow better convergence. 

\subsection{Trade-off between accuracy and skip ratio}
ActionSpotter manages the lack of spot frame ground truth and tackles the spotting problem by optimizing the mAP. 
Moreover, it allows to balance the mAP  and the browser skip ratio.
We report in Table \ref{tab:speed} the mAP and the skip ratio for different values of $\gamma$ to highlight the impact of the discount factor on the policy learning.

\begin{table}[t]
\begin{center}
\begin{tabular}{|l|c|c|c|c|c|}
\hline
\multirow{2}{*}{Our}& \multicolumn{5}{c|}{ $\gamma$}\\\cline{2-6}
 & 1.0 & 0.99 & 0.98 & 0.96 & 0.95\\
\hline\hline
mAP (\%) & \textbf{65.6} & 64.3 & 63.4 & 61.4 & 61.3\\
skip ratio (\%)  & 23  & 29 & \textbf{53} & 51 & 51\\
\hline
\end{tabular}
\end{center}
\caption{\label{tab:speed} \textbf{Performance on THUMOS14 validation set according to the discount factor $\gamma$}.}
\end{table}

As expected, for small values, the lower the $\gamma$ discount factor, the higher the skip ratio (convergence issues appear quickly when the $\gamma$ decreases).
But, as a trade-off exists between speed and accuracy, accuracy decreases too. 
Nevertheless, when the skip ratio increases from 23\% to 53\%, the mAP decreases only by 2\%. 
For $\gamma=0.98$, our algorithm is still better than the best detector for spotting while using only 47\% of the frames.
Currently, with the presented action space setting that puts a lot of emphasis on mAP, it is difficult to go below a 53\% skip ratio. But, by adding more actions, it is possible to skip more frames even if mAP decreases significantly.
For example, a mAP of 50.9\% is obtained using only 2\% of the frames (same skip ratio as \cite{yeung2016end}). 
But there is no point in skipping so many frames if it leads to such low performance level. As mentioned before, it seems difficult to perform accurate detection or spotting with 2\% of frames, unlike classification \cite{wu2019adaframe} which assumes only one action.

Thus, while skipping many frames degrades performance, skipping some frames improves it.
Indeed, by removing the browser and processing all video frames, the mAP decreases from 65.6\% to 64.0\%. Using the browser agent leads to a 1.6\% improvement in mAP while using 23\% fewer frames. 

\begin{figure}[h]
\centering
\includegraphics[width=0.42\textwidth]{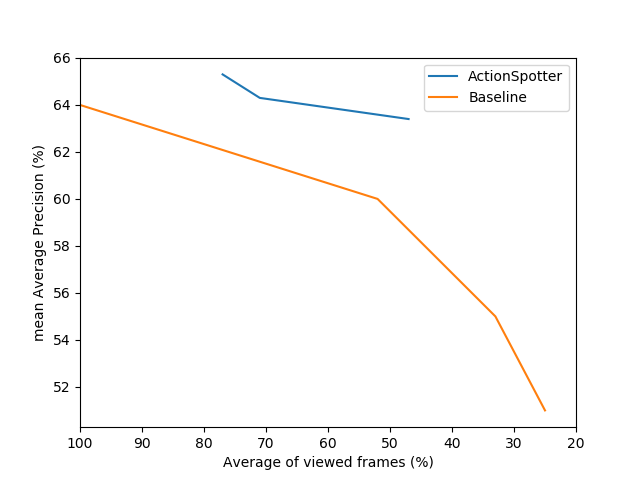}
\caption{\textbf{Benefit of reinforcement browsing}: mAP according to the percentage of viewed frames for ActionSpotter algorithm and uniform sub-sampling.
\label{fig:curvemapskip}}
\end{figure}

It is interesting to note that we also train our pipeline without the Browser but with uniform subsampling. Precisely, we subsample videos with different sampling rates and observe the mAP.  
Figure \ref{fig:curvemapskip} shows that ActionSpotter produces better results regardless of the skip ratio.

\section{Conclusion}
We propose an algorithm called ActionSpotter able to tackle action spotting: collecting one frame per action instance.
Our algorithm is based on state-of-the-art reinforcement algorithm and is able to tackle action spotting in a streaming context (frames are not stored) while skipping some video frames.

To evaluate the proposed algorithm, we introduce a metric that quantifies the accuracy of the action spotting. It is based on the mean Average Precision (mAP) conventionally used in detection.
The main result of this paper is that the adaptation of state-of-the-art action detectors into spotting algorithms is less efficient that learning an end-to-end spotting: ActionSpotter reaches 65\% of mAP (while skipping 23\% of video frames), when state-of-the-art detectors only reach 59\% of mAP (with a dense exploration).

Indeed, unlike action detectors that have to deal with ambiguous temporal boundaries, ActionSpotter focuses on  extracting one frame per instance thanks to reinforcement learning. 

\section*{Acknowledgments}
This work was performed using HPC resources from GENCI-IDRIS (2019-AD011011269)





%


\bibliographystyle{IEEEtran}
\bibliography{IEEEabrv,egbib}

%
%

\end{document}